\documentclass[conference]{IEEEtran}
\IEEEoverridecommandlockouts
\usepackage{cite}
\usepackage{amsmath,amssymb,amsfonts}
\usepackage{algorithmic}
\usepackage{graphicx}
\usepackage{textcomp}
\usepackage{xcolor}

\usepackage{multirow}

\usepackage{booktabs}
\usepackage{graphicx}
\usepackage{hyperref}

\def\BibTeX{{\rm B\kern-.05em{\sc i\kern-.025em b}\kern-.08em
    T\kern-.1667em\lower.7ex\hbox{E}\kern-.125emX}}
\begin{document}

\title{Complex Network for Complex Problems: A comparative study of CNN and Complex-valued CNN\\
}


\author{\IEEEauthorblockN{Soumick~Chatterjee\textsuperscript{\textsection}\IEEEauthorrefmark{1}\IEEEauthorrefmark{2}\IEEEauthorrefmark{3}, 
                         Pavan Tummala\textsuperscript{\textsection}\IEEEauthorrefmark{1}, Oliver~Speck\IEEEauthorrefmark{3}\IEEEauthorrefmark{4}\IEEEauthorrefmark{5} and  
                         Andreas~N{\"u}rnberger\IEEEauthorrefmark{1}\IEEEauthorrefmark{2}\IEEEauthorrefmark{5}
                         }
\\
\IEEEauthorblockA{\IEEEauthorrefmark{1}Faculty of Computer Science, Otto von Guericke University Magdeburg, Germany}
\IEEEauthorblockA{\IEEEauthorrefmark{2}Data and Knowledge Engineering Group, Otto von Guericke University Magdeburg, Germany}
\IEEEauthorblockA{\IEEEauthorrefmark{3}Biomedical Magnetic Resonance, Otto von Guericke University Magdeburg, Germany}
\IEEEauthorblockA{\IEEEauthorrefmark{4}German Center for Neurodegenerative Disease, Magdeburg, Germany}
\IEEEauthorblockA{\IEEEauthorrefmark{5}Center for Behavioral Brain Sciences, Magdeburg, Germany}
}
\maketitle

\begingroup\renewcommand\thefootnote{\textsection}
\footnotetext{S. Chatterjee and P. Tummala contributed equally}
\endgroup

\begin{abstract}
Neural networks, especially convolutional neural networks (CNN), are one of the most common tools these days used in computer vision. Most of these networks work with real-valued data using real-valued features. Complex-valued convolutional neural networks (CV-CNN) can preserve the algebraic structure of complex-valued input data and have the potential to learn more complex relationships between the input and the ground-truth. Although some comparisons of CNNs and CV-CNNs for different tasks have been performed in the past, a large-scale investigation comparing different models operating on different tasks has not been conducted. Furthermore, because complex features contain both real and imaginary components, CV-CNNs have double the number of trainable parameters as real-valued CNNs in terms of the actual number of trainable parameters. Whether or not the improvements in performance with CV-CNN observed in the past have been because of the complex features or just because of having double the number of trainable parameters has not yet been explored. This paper presents a comparative study of CNN, CNNx2 (CNN with double the number of trainable parameters as the CNN), and CV-CNN. The experiments were performed using seven models for two different tasks - brain tumour classification and segmentation in brain MRIs. The results have revealed that the CV-CNN models outperformed the CNN and CNNx2 models.
\end{abstract}


\section{Introduction}
Convolutional neural networks (CNN) have become one of the most predominant parts of the field of computer vision. Since its inception, CNNs have been widely used for different applications - image classification, segmentation, anomaly detection, etc. Several network architectures have been proposed for different tasks. For example, ResNet~\cite{he2016deep} for classification or UNet~\cite{ronneberger2015u} for segmentation and inverse problems like reconstruction, and their variations have become widely popular. The main building block of CNNs is the convolution operation, which is calculated by the integral of the product of two functions - the input ($x$) and the kernel ($w$), and the output is referred to as the feature map or activation map ($s$), and is given by:  
\begin{equation}\label{eq:conv1D}
    s(t)=(w \star x)(t)=\sum_{a} x(t+a) w(a) 
\end{equation}
Here both $w$ and $x$ are real-valued. Complex-valued convolutional networks~\cite{trabelsi2018deep} or CV-CNNs extend this with the help of complex-valued convolution operation that can be defined by:
\begin{equation}\label{eq:complexconv}
\begin{aligned}
     s_r(t)=(w_r \star x_r)(t) - (w_i \star x_i)(t)\\
     s_i(t)=(w_i \star x_r)(t) + (w_r \star x_i)(t)
\end{aligned}
\end{equation}
where $x_r$ and $x_i$ are the real and imaginary parts of the complex-valued input $x$. Similarly, $w_r$ and $w_i$ are of the complex-valued kernel $w$, and $s_r$ and $s_i$ are of the generated complex-valued feature map $s$. This can also be written using matrix notation as:
\begin{equation}\label{eq:complexconv_matrix}
\left[\begin{array}{c}
\Re(\mathbf{w} \star \mathbf{x}) \\
\Im(\mathbf{w} \star \mathbf{x})
\end{array}\right]=\left[\begin{array}{cr}
\mathbf{w_r} & -\mathbf{w_i} \\
\mathbf{w_i} & \mathbf{w_r}
\end{array}\right] \star\left[\begin{array}{l}
\mathbf{x_r} \\
\mathbf{x_i}
\end{array}\right]
\end{equation}
CV-CNNs can preserve the algebraic structure of complex-valued data and also have the potential to learn more intricate representations. In many cases, CV-CNNs~\cite{trabelsi2018deep} outperformed regular convolutional networks (CNNs)~\cite{choi2018phase,zhang2020cv,ziller2021complex,mu2021cv}. Nonetheless, it has not been widely employed, which might be related to the fact that complex-valued convolution technology is less established than real-valued convolution technology~\cite{kim2019loraki}. The addition of complex-valued convolution to the PyTorch framework~\cite{pytorch_NEURIPS2019_9015} expanded the possibilities of building such techniques.

Trabelsi et al.~\cite{trabelsi2018deep} demonstrated the superiority of CV-CNNs, even when the input is only real-valued. The raw data acquired in magnetic resonance imaging (MRI), known as the k-space, is complex-valued. However, most of the approaches proposed for the reconstruction of undersampled MRIs working in the k-space~\cite{akccakaya2019scan,han2019k,du2021adaptive}, or even methods based on a combination of the reconstructed image space and k-space~\cite{zhu2018image,hammernik2018learning,sriram2020end}, treat complex-valued input as two separate real-valued data by splitting the complex data into real and imaginary parts and then concatenating them on the channel dimension to supply to real-valued neural network models - ignoring the complex data's richer algebraic structure~\cite{cole2021analysis}. \cite{mickISMRM21ksp} showed the potential of using CV-CNNs for undersampled MRI reconstruction, but does not provide a comparison against real-valued CNNs.

Even though some comparative studies of CNNs and CV-CNNs have been performed for different tasks, a wide-scale study comparing different models working on different tasks has not been performed. Moreover, another thing to note is that in terms of the actual number of trainable parameters, CV-CNNs have double the number of trainable parameters as regular CNNs, as the complex features (kernels) have real and imaginary parts. This leads to the question of whether the improvements observed in the previous studies were because of this increase in the number of trainable parameters or really because of the complex networks - this has not yet been answered. Hence, this research focuses on comparing several CNN models against identical CV-CNN models (convolutions being replaced with complex-valued convolution) and against the same CNN models with double the number of parameters.


\section{Methodology}
This study presents a comparison of CNNs with CV-CNNs for classification and segmentation tasks where each pair of networks (CNN and CV-CNN) has the same amount of parameters (features). The sole distinction was that CNNs used real-valued features (kernels), whereas CV-CNNs used complex-valued ones. However, in terms of the precise number of trainable parameters, the CV-CNNs had twice as many as the CNNs since they were complex-valued - each of them had real and imaginary components. Hence, to be fair in terms of the number of trainable parameters, one more set of networks was added in this research, which was real-valued CNNs, but with twice the number of trainable parameters as their CNN counterparts - having the same number of trainable parameters as the CV-CNN, but twice the number of features as both CNN and CV-CNN (referred here as CNNx2). This increase in terms of the number of trainable parameters was performed by increasing the number of features in each layer of those network architectures. To facilitate the implementation of the CV-CNNs with minimum efforts, a wrapper around the PyTorch layers as a python package pytorch-complex~\footnote{pytorch-complex package: \url{https://github.com/soumickmj/pytorch-complex}} has been developed during this research.

\subsection{Network Models}
\label{sec:recon_mixspace_cnnVcvcnn_models}
Three different complexity levels of the ResNet~\cite{he2016deep} model was used in this study for the task of image classification. The original torchvision implementation of ResNet was used as the CNN models. They were then modified in two ways: the number of features for all layers was increased by a factor of $2$ to create the CNNx2 models, and all layers of the implementation were changed into complex-valued layers using the pytorch-complex package to create the CV-CNN models. Four different models were utilised for image segmentation: UNet~\cite{ronneberger2015u}, Attention UNet~\cite{Oktay2018}, a modified version of ReconResNet~\cite{chatterjee2021reconresnet}, and TransUNet~\cite{chen2021transunet}. These CNN models were modified in the same way as the classification models were to create the CNNx2 and CV-CNN models. Table~\ref{tab:params} presents the number of trainable parameters (as well as features) present in the aforementioned classification and segmentation CNN models. The CNNx2 models had exactly double these numbers in terms of features and trainable parameters, while the CV-CNN models had the same number of features but double the number of trainable parameters. 

\begin{table}[]
\centering
\caption{The number of features, as well as trainable parameters, present in the chosen CNN models}
\label{tab:params}
\resizebox{0.4\textwidth}{!}{%
\begin{tabular}{@{}cc@{}}
\toprule
\textbf{Network Architecture} & \textbf{\begin{tabular}[c]{@{}c@{}}Number of Trainable \\ Parameters and Features\end{tabular}} \\ \midrule
ResNet18       & 11.4M \\
ResNet34       & 21.5M \\
ResNet50       & 23.9M \\ \midrule
ReconResNet    & 17.3M \\
UNet           & 31.4M \\
Attention UNet & 34.3M \\
TransUNet      & 43.5M \\ \bottomrule
\end{tabular}%
}
\end{table}

\subsection{Dataset}
\label{sec:recon_mixspace_cnnVcvcnn_dataset_prelim}
The BraTS 2020 dataset~\cite{menze2014multimodal,lloyd2017high,bakas2018identifying} was used in this study, and it contains multimodal MRIs: T1-weighted (T1w), post-contrast T1-weighted (T1ce), T2-weighted (T2), and T2-FLAIR (or simply, FLAIR) of 346 patients with two different types of tumours: low-grade glioma (LGG, 73 patients) and high-grade glioma or glioblastoma (HGG, 273 patients). The collection includes 3D volumes. The tumour-free slices were used to form a third class, which resulted in $4,731$, $18,276$, and $24,679$ slices for the LGG, HGG, and tumour-free classes, respectively - for a 3-class classification task. For the task of image segmentation, the models attempted to segment the tumour (both LGG and HGG) as a binary problem. In both cases, only the T1ce MRIs were used as input. This dataset only contains real-valued magnitudes images. As a result, even though the input to the models was complex-valued, there was no phase in the input.

\subsection{Implementation, Training and Evaluation}
\label{sec:recon_mixspace_cnnVcvcnn_implement}
The loss during the task of image classification was calculated using the cross-entropy loss, while for the task of image segmentation, Dice loss was employed. All the models were trained to convergence with the Adam optimiser for 80 epochs with a batch size of 64 with a learning rate of 1e-4. Before slicing in 2D and supplying to the network models, the 3D volumes were interpolated to have isotropic voxels of 1mm and the intensity values were normalised. Additionally, data augmentation techniques using TorchIO~\cite{perez2021torchio} were used during training. Two classes of data augmentations were applied - intensity augmentation that included bias field artefacts, Gaussian noise, random gamma manipulation, ghosting artefacts, and spatial augmentation that included horizontal flipping, vertical flipping, affine transformations, rotation. The classification results were evaluated using the F1-score and accuracy, while the segmentation performance was evaluated using the Dice score and intersection over union (IoU). 5-fold cross-validation was performed for evaluating the classification models by randomly dividing the dataset into five subsets five times - each time using four subsets for training and one for testing. Similar for the segmentation models, 3-fold cross-validation was performed. 

\section{Results and Discussion}
\label{sec:recon_mixspace_cnnVcvcnn_results_prelim}
\begin{figure*}[htbp]
\centerline{\includegraphics[width=\textwidth]{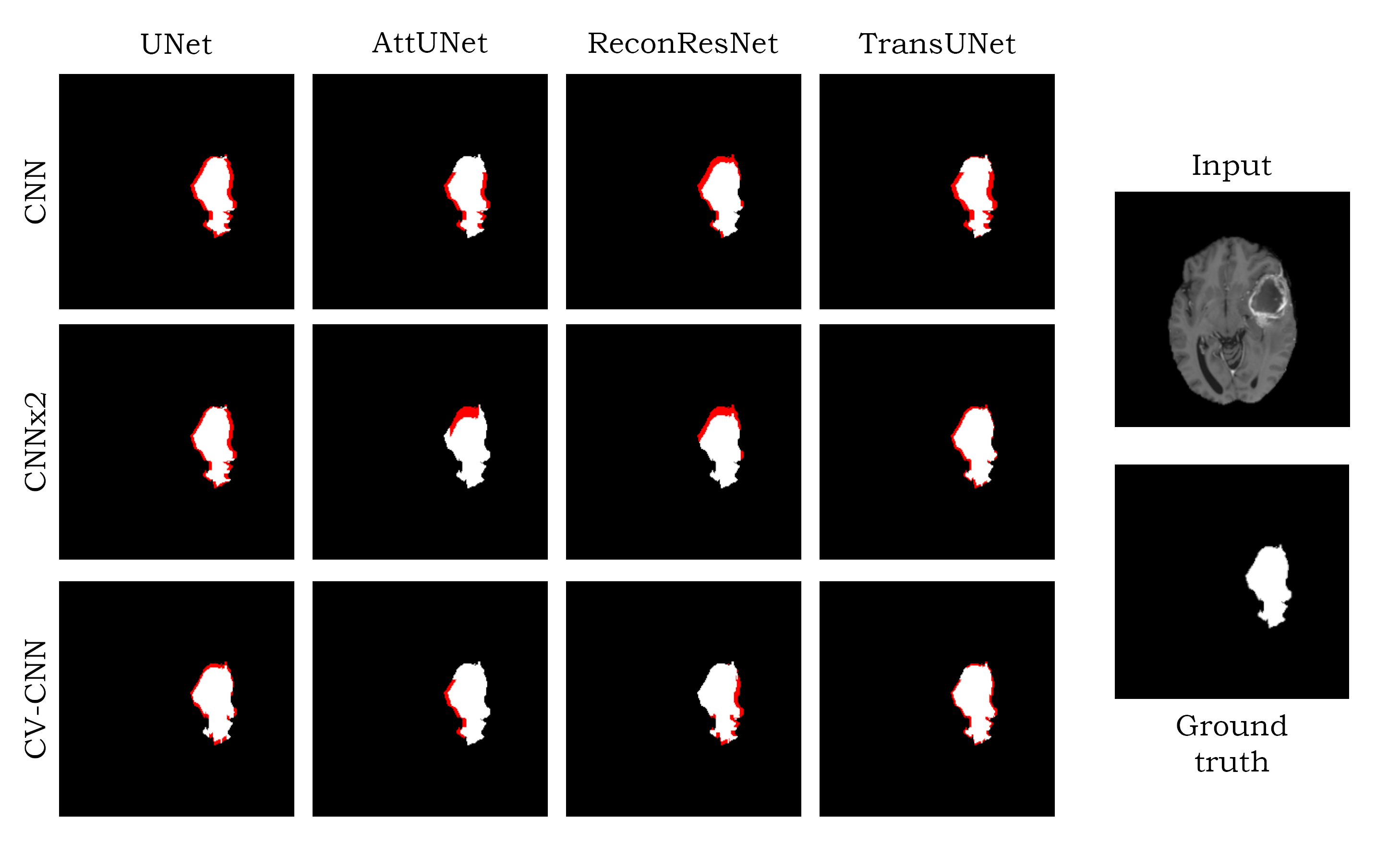}}
\caption{Segmentation predictions of different models for an example slice (T1ce MRI): red denotes under-segmentation}
\label{fig:seg}
\end{figure*}

Tables~\ref{tab:cnnVcvcnn_classify}~and~\ref{tab:cnnVcvcnn_seg} present the comparative results of the CNN and CV-CNN models for classification and segmentation tasks, respectively. The ResNet18 CV-CNN model can out as the best performing model during classification, with an f1-score of 0.829±0.119. While segmenting tumours, the CV-CNN version of the attention UNet resulted in the best scores, including a Dice score of 0.868±0.089. In terms of both f1-score and accuracy, CNNx2 performed better than the CNNs, while CV-CNN outperformed them both. When it comes to segmentation, the same trend regarding CNN and CNNx2 can be observed. CV-CNN models outperformed the other two in terms of IoU for all four architectures, while in terms of the Dice score for TransUNet, both CNNx2 and TransUNet resulted in the same mean score, while CNNx2 had less standard deviation. Moreover, Fig.~\ref{fig:seg} portrays the results from the different segmentation models for an example slice. It can be observed that CV-CNN models resulted in less under-segmentation compared to the other ones.

Two general conclusions may be drawn from the results. The CNNx2 models outperformed the CNN models, which may be attributed to the fact that the CNNx2 models contained twice as many features as the CNN models. The second observation is that the CV-CNN models outperformed both the CNN and CNNx2 models, despite having the same number of features as the CNN models (but twice the number of trainable parameters) and half the number of features as the CNNx2 models (but having the same number of trainable parameters). These tests convincingly demonstrate that the advantages are due to the complex-valued features rather than the CV-CNN models having a twofold number of trainable parameters when the number of features is the same for both.

The results revealed ResNet18 as the winner for the classification task despite having the least number of trainable parameters. This might be because, as a smaller, less complicated model, it was less prone to overfitting, and the number of features in ResNet18 may have been sufficient to train for this specific problem. Even though Attention UNet beat the other three models in the segmentation task, one intriguing discovery was made about the ReconResNet model, which was originally proposed for the job of artefact removal in undersampled MRIs~\cite{chatterjee2021reconresnet}. This model also has the least number of trainable parameters (Table~\ref{tab:params}) among the segmentation models. The results reveal that this model can also do segmentations, performing better than the UNet and TransUNet models, which were both originally proposed for the task of image segmentation.

\begin{table}[]
\centering
\caption[CNN and CV-CNN: Classification results]{Classification results of CNN and CV-CNN}
\label{tab:cnnVcvcnn_classify}
\resizebox{0.49\textwidth}{!}{%
\begin{tabular}{@{}cccc@{}}
\toprule
Model                     & Type            & F1-Score                   & Accuracy                   \\ \midrule
\multirow{3}{*}{ResNet18} & CNN             & 0.819 $\pm$ 0.101          & 0.809 $\pm$ 0.056          \\
                          & CNNx2           & 0.827 $\pm$ 0.097          & 0.811 $\pm$ 0.121          \\
                          & \textbf{CV-CNN} & \textit{\textbf{0.829 $\pm$ 0.119}} & \textit{\textbf{0.835 $\pm$ 0.129}} \\
\multirow{3}{*}{ResNet34} & CNN             & 0.783 $\pm$ 0.107          & 0.798 $\pm$ 0.115          \\
                          & CNNx2           & 0.798 $\pm$ 0.085          & 0.805 $\pm$ 0.121          \\
                          & \textbf{CV-CNN} & \textbf{0.817 $\pm$ 0.072} & \textbf{0.829 $\pm$ 0.123} \\
\multirow{3}{*}{ResNet50} & CNN             & 0.790 $\pm$ 0.126          & 0.788 $\pm$ 0.125          \\
                          & CNNx2           & 0.796 $\pm$ 0.089          & 0.795 $\pm$ 0.118          \\
                          & \textbf{CV-CNN} & \textbf{0.807 $\pm$ 0.115} & \textbf{0.819 $\pm$ 0.097} \\ \bottomrule
\end{tabular}%
}
\end{table}

\begin{table}[]
\centering
\caption[CNN and CV-CNN: Segmentation results]{Segmentation results of CNN and CV-CNN}
\label{tab:cnnVcvcnn_seg}
\resizebox{0.49\textwidth}{!}{%
\begin{tabular}{@{}cccc@{}}
\toprule
Model                        & Type            & Dice                       & IoU                        \\ \midrule
\multirow{3}{*}{UNet}        & CNN             & 0.752 $\pm$ 0.011          & 0.679 $\pm$ 0.121          \\
                             & CNNx2           & 0.759 $\pm$ 0.193          & 0.681 $\pm$ 0.089          \\
                             & \textbf{CV-CNN} & \textbf{0.789 $\pm$ 0.106} & \textbf{0.709 $\pm$ 0.112} \\
\multirow{3}{*}{\begin{tabular}[c]{@{}c@{}}Attention\\ UNet\end{tabular}} & CNN             & 0.848 $\pm$ 0.070                   & 0.733 $\pm$ 0.111                   \\
                             & CNNx2           & 0.849 $\pm$ 0.103          & 0.742 $\pm$ 0.087          \\
                                                                          & \textbf{CV-CNN} & \textit{\textbf{0.868 $\pm$ 0.089}} & \textit{\textbf{0.771 $\pm$ 0.105}} \\
\multirow{3}{*}{ReconResNet} & CNN             & 0.829 $\pm$ 0.139          & 0.743 $\pm$ 0.098          \\
                             & CNNx2           & 0.831 $\pm$ 0.112          & 0.755 $\pm$ 0.108          \\
                             & \textbf{CV-CNN} & \textbf{0.853 $\pm$ 0.540}  & \textbf{0.768 $\pm$ 0.122} \\ 
\multirow{3}{*}{TransUNet} & CNN             & 0.841 $\pm$ 0.107          & 0.741 $\pm$ 0.074          \\
                             & CNNx2           & \textbf{0.847} $\pm$ \textbf{0.137}          & 0.747 $\pm$ 0.113          \\
                             & \textbf{CV-CNN} & \textbf{0.847} $\pm$ 0.219  & \textbf{0.749 $\pm$ 0.172} \\ \bottomrule
\end{tabular}%
}
\end{table}

\section{Conclusion and Future Work}
This research presented a comprehensive comparative study of real-valued CNN versus complex-valued CNN for brain tumour classification and segmentation. For the same, seven different network architectures which were proposed using real-valued convolutions were modified with complex-valued convolutions and presented the CV-CNN versions of those networks that have not been proposed before. The results of the experiments show that CV-CNN models outperform identical real-valued CNN models with the same number of features and models with the same number of trainable parameters. This research demonstrates the superiority of CV-CNNs over CNNs because of the complex-valued features and not only because of having more parameters to learn. However, this research only uses one dataset for this task. An extensive study involving different problem statements and different datasets will be conducted in the near future to conclusively prove the superiority of the CV-CNN over CNN. While working with complex-valued input images (not only the magnitude MRIs), these CV-CNNs might be a better choice. With the recent release of pathological annotations for the fastMRI dataset~\cite{zbontar2018fastmri}, known as the fastMRI+~\cite{zhao2022fastmriplus}, this CNN vs CV-CNN research may be greatly expanded by incorporating complex-valued data rather than merely magnitude data, as was done here. Furthermore, because fastMRI contains raw data, one intriguing future avenue to pursue will be to apply CV-CNN classification models directly to the raw k-space data and investigate the feasibility of classifying directly in the k-space.


\bibliographystyle{IEEEtran}  
\bibliography{ref}

\end{document}